%% file: neurips_2023.tex
\documentclass{article}

\usepackage[preprint]{neurips_2023}

\usepackage[utf8]{inputenc} 
\usepackage[T1]{fontenc}    
\usepackage{url}            
\usepackage{booktabs}       
\usepackage{amsfonts}       
\usepackage{nicefrac}       
\usepackage{microtype}      
\usepackage{xcolor}         
\usepackage{makecell}  

\definecolor{linkColor}{rgb}{0.18,0.39,0.62}
\usepackage[colorlinks=true,linkcolor=linkColor,citecolor=linkColor,filecolor=linkColor,urlcolor=linkColor]{hyperref}       
\usepackage[most]{tcolorbox}
\usepackage{enumitem}
\usepackage{amsmath}
\usepackage{amssymb}
\usepackage{mathtools}
\usepackage{amsthm}
\usepackage{graphicx}
\usepackage{xspace}
\usepackage{pifont}
\usepackage{array}
\usepackage{bm}
\usepackage{ulem}
\usepackage{multirow}

\usepackage{subcaption}
\usepackage{tablefootnote}
\usepackage{arydshln}
\usepackage{adjustbox}

\makeatletter

\newcommand{\Rmnum}[1]{\expandafter\@slowromancap\romannumeral #1@}
\makeatother

\title{Boosting Large Language Model for Speech Synthesis: An Empirical Study}

\author{%
 Hongkun Hao$^1$\thanks{Work was done during internship at Microsoft Research Asia.}, \ Long Zhou$^2$, \ Shujie Liu$^2$, \ Jinyu Li$^2$, \ Shujie Hu$^2$, \ Rui Wang$^1$,  \ Furu Wei$^2$ \\
$^1$Shanghai Jiao Tong University\\
$^2$Microsoft Corporation
}

\begin{document}

\maketitle

\begin{abstract}
Large language models (LLMs) have made significant advancements in natural language processing and are concurrently extending the language ability to other modalities, such as speech and vision. Nevertheless, most of the previous work focuses on prompting LLMs with perception abilities like auditory comprehension, and the effective approach for augmenting LLMs with speech synthesis capabilities remains ambiguous. In this paper, we conduct a comprehensive empirical exploration of boosting LLMs with the ability to generate speech, by combining pre-trained LLM LLaMA/OPT and text-to-speech synthesis model VALL-E. We compare three integration methods between LLMs and speech synthesis models, including directly fine-tuned LLMs, superposed layers of LLMs and VALL-E, and coupled LLMs and VALL-E using LLMs as a powerful text encoder. Experimental results show that, using LoRA method to fine-tune LLMs directly to boost the speech synthesis capability does not work well, and superposed LLMs and VALL-E can improve the quality of generated speech both in speaker similarity and word error rate (WER). Among these three methods, coupled methods leveraging LLMs as the text encoder can achieve the best performance, making it outperform original speech synthesis models with a consistently better speaker similarity and a significant (10.9\%) WER reduction. 
\end{abstract}

\section{Introduction}
\setcounter{footnote}{0}

The emergence of large language models (LLMs), such as ChatGPT \citep{OpenAI2023GPT4TR} and LLaMA \citep{touvron2023llama}, has revolutionized most traditional natural language processing (NLP) tasks, like text summarization and dialogue system. The powerful language generation capabilities of LLMs have prompted exploration into their applications in other modalities, e.g., speech and vision \citep{OpenAI2023GPT4V,huang2023language,zhang2023video}.
For example, GPT-4V \citep{OpenAI2023GPT4V} enables users to instruct GPT-4 to analyze image inputs they provided. Video-LLaMA \citep{zhang2023video} empowers LLM with the ability to comprehend both visual and auditory content present in video.
These multi-modal LLMs provide the potential to enhance the impact of text-only systems by integrating new interfaces and functionalities, allowing them to handle new tasks and deliver fresh experiences to users.

Regarding the application of LLMs to speech, the majority of earlier research primarily concentrates on aligning speech representation with the LLM input space \citep{wu2023decoder,fathullah2023prompting,shu2023llasm,tang2023salmonn}.
For instance, Speech-LLaMA \citep{wu2023decoder} proposes an effective method to accomplish speech-to-text tasks by leveraging Connectionist Temporal Classification (CTC) \citep{graves2006connectionist} model and audio encoder to map the compressed acoustic features to the continuous semantic space of the LLM.
LLaSM \citep{shu2023llasm} takes advantage of a well-trained Whisper encoder to encode the speech signals into hidden states, and utilizes a modal adaptor to align the above output hidden states with the input text embedding of LLMs.
Compared to understanding speech, enabling LLMs to generate speech is considerably more challenging, given that speech is a continuous signal significantly deviating from the output space of LLMs.
To enable speech generation ability, existing works such as SpeechGPT \citep{zhang2023speechgpt} and AudioPaLM \citep{rubenstein2023audiopalm} employ the approach of directly fine-tuning a pre-trained LLM, which requires substantial computational resources and time.
How to effectively enhance LLMs with the capabilities for speech synthesis remains a relatively unexplored area. 

\begin{figure*}[htbp]
	\centering
	\begin{subfigure}{0.49\linewidth}
		\centering
		\includegraphics[width=\linewidth]{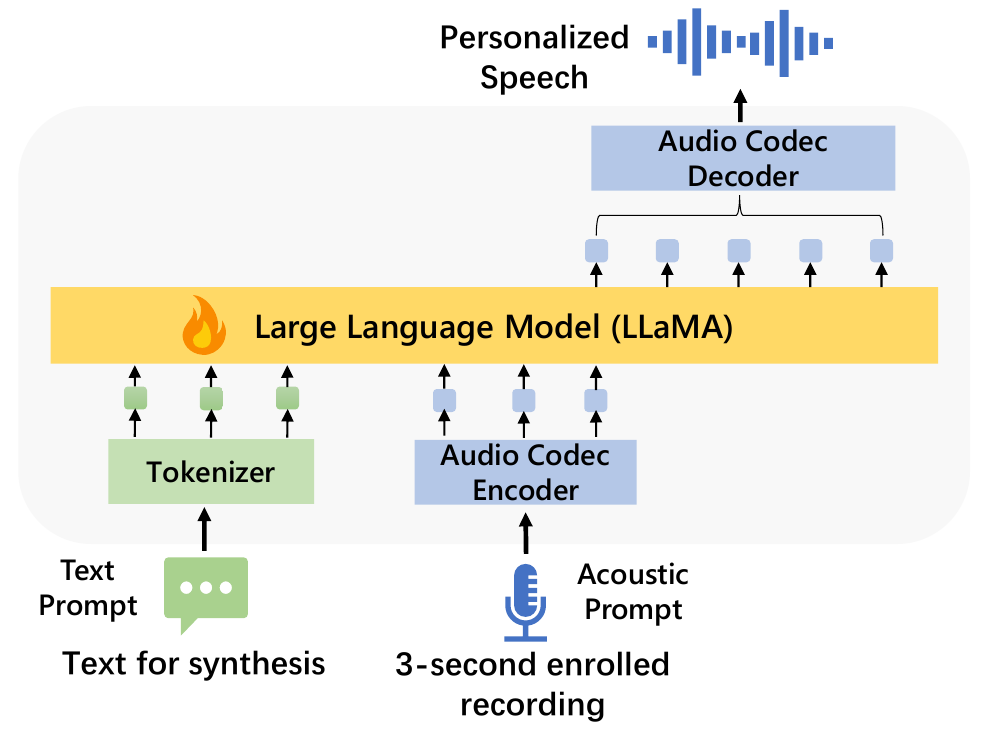}
		\caption{Method A: Directly Fine-tuned LLM}
		\label{fig:methodA}
        \vspace{5mm}
	\end{subfigure}
	\centering
	\begin{subfigure}{0.49\linewidth}
		\centering
		\includegraphics[width=\linewidth]{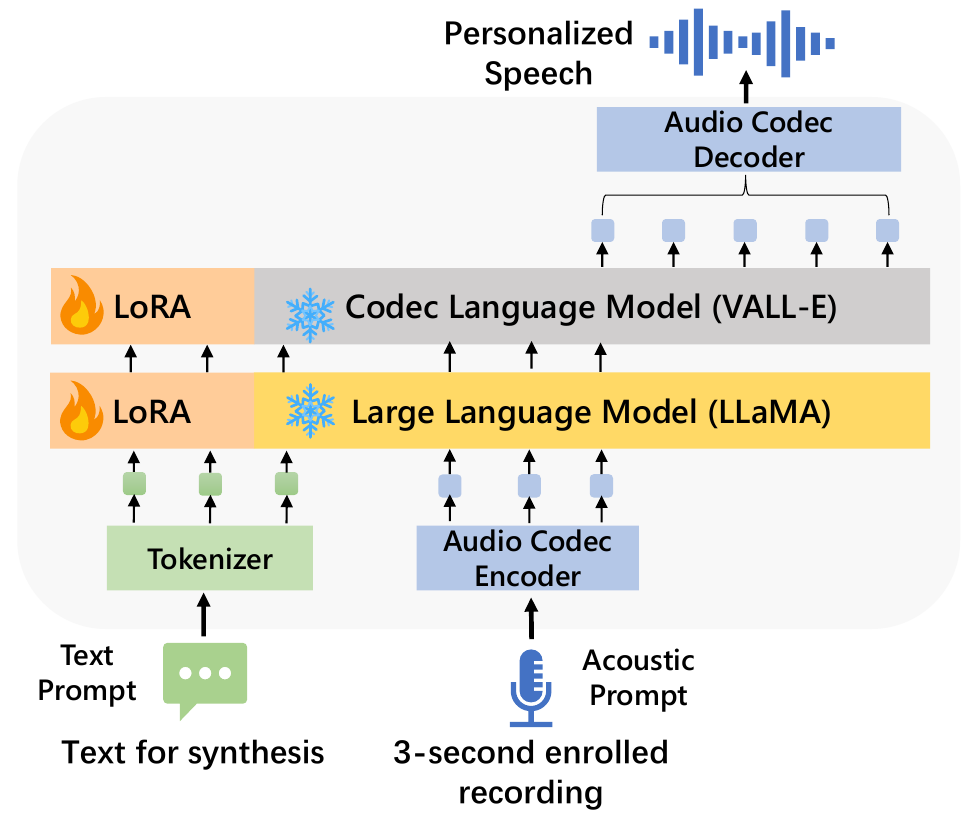}
		\caption{Method B: Superposed LLM and VALL-E}
		\label{fig:methodB}
        \vspace{5mm}
	\end{subfigure}
    \begin{subfigure}{0.80\linewidth}
    \centering
    \includegraphics[width=0.99\linewidth]{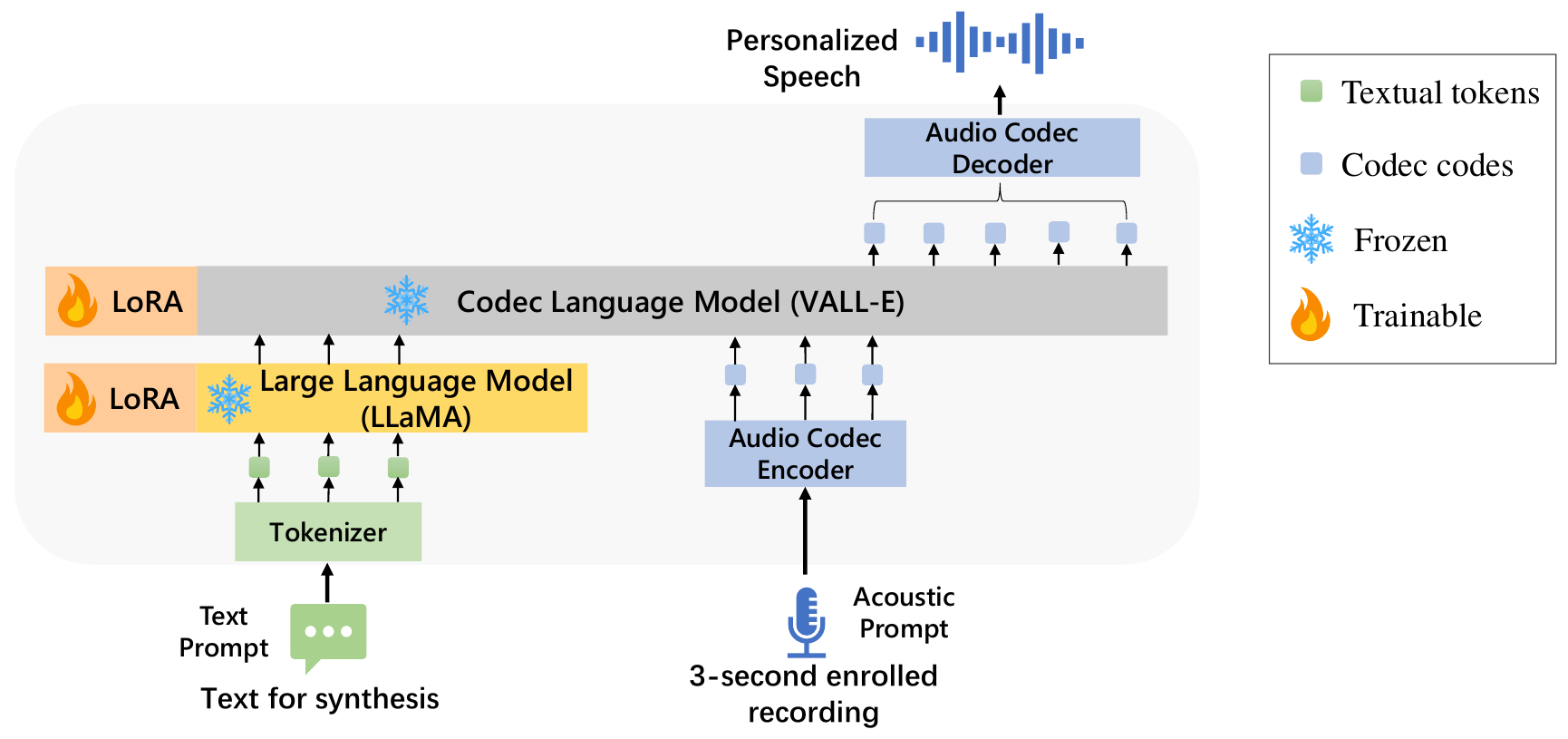}
    \caption{Method C: Coupled LLM and VALL-E}
    \label{fig:methodC}\
	\end{subfigure}
    \caption{Overview of the proposed different integration methods.
    (a) Method A: Directly fine-tuned LLMs where LLMs are trained for predicting codec codes with an expanded vocabulary. (b) Method B: Superposed LLMs and VALL-E, where both LLMs and VALL-E are used to model textual tokens and acoustic tokens successively. (c) Method C: Coupled LLMs and VALL-E, where the better text representation provided by LLM is regarded as the textual input of VALL-E.
    }
	\label{fig:methods}
\end{figure*}

To better understand this task, we are going to answer two questions: 1) Can the codec codes be treated by LLMs simply as a kind of language similar to other natural languages? 2) What kind of information can LLMs provide to improve the quality of synthesized speech? 
In order to answer these two questions, in this paper, we propose and compare several integration approaches to enable the LLMs with speech synthesis capability. 
In this study, we focus on zero-shot text-to-speech (TTS) tasks following the state-of-the-art model VALL-E \citep{wang2023neural_valle}, which mainly uses an auto-regressive (AR) Transformer decoder model to predict the discrete token of speech depending on the corresponding textual tokens.
To enhance the speech generation of LLMs, we first discretize the continuous speech into multi-layer discrete codec codes via audio compression model Encodec \citep{defossez2022high}, and expand the vocabulary of LLMs with the vocabulary of codec codes, e.g., 1024 tokens. 
We design three combination strategies to achieve the first-layer codec code prediction with LLM, like the AR model in VALL-E, as follows:

\begin{itemize}
\item \textbf{Directly Fine-tuned LLMs}. We directly fine-tune large language models via paired text and codec codes from speech recognition dataset, with full parameters or partial parameters (LoRA \citep{hu2021lora}), as shown in Figure \ref{fig:methods}(\subref{fig:methodA}).
\item \textbf{Superposed LLMs and VALL-E}. Figure \ref{fig:methods}(\subref{fig:methodB}) illustrates this strategy that we superimpose the two models into one model. In this method, we use the large language model to encode both textual tokens and acoustic tokens, and then we feed them into the codec language model VALL-E.
\item \textbf{Coupled LLMs and VALL-E}. As shown in Figure \ref{fig:methods}(\subref{fig:methodC}), we use an additional text-based large language model to encode the text sequence and then input them into the VALL-E AR model. 
The coupled method differs from the aforementioned superposed approach as it does not utilize LLMs to model codec codes.
\end{itemize}

After that, we can use the non-autoregressive (NAR) model of VALL-E to generate the codec codes of the rest quantizers, and utilize the Encodec decoder to recover the waveform of the speech.
Models are trained on 44.5K hours Multilingual Librispeech English data and 960 hours LibriSpeech data and evaluated on LibriSpeech dev-clean, dev-other, test-clean, and test-other datasets.
Experimental results demonstrate that coupled LLMs and VALL-E can achieve the best performance among baseline and our methods. 
Additionally, we perform thorough analyses of various facets of our approach, examining the impact of model size, the benefits of continuous pre-training, the effect of the pre-trained VALL-E, and a comparative evaluation of LoRA versus complete fine-tuning for VALL-E.
Based on the results, we can draw conclusions as follows:
\begin{itemize}
\item  Codec codes can not be simply treated as another language since the results of directly fine-tuned LLM are not promising. The reason could be that, the sequence length of codec codes is much longer than the length of corresponding text, and also the information provided by codec codes is much more fine-grained and more diverse than that of text. 
\item While LLMs with LoRA may not excel at generating codec codes, they can serve as a unified encoder for processing both text and codec codes. The outputs generated by LLMs can provide valuable representation for a codec language model (e.g., VALL-E) to produce more accurate codec codes.
\item LLM can be used as a powerful text encoder alone that can model pertinent and extensive content information, which is instrumental for VALL-E to generate speech of superior quality and enhanced robustness.
The structure using LLM as a text encoder, coupled with a dedicated decoder module such as VALL-E, achieves the best performance.
\end{itemize}




\section{Related Work} \label{related_work}



Our work is typically based on LLMs, which have made significant breakthroughs in natural language processing, outperform previous state-of-the-art models in extensive NLP tasks, and inspire the instruction-following capability to achieve the unseen tasks \citep{ouyang2022training, OpenAI2023GPT4TR, touvron2023llama, anil2023palm}. The advent of ChatGPT \citep{ouyang2022training} marks a transformative era in the field of artificial intelligence. By leveraging vast training datasets and extensive parameter configurations, in conjunction with instruction tuning and RLHF algorithm, it raises amazing emergent abilities and becomes the best artificial intelligence assistant with natural language as an interface.

Given that the world's information not only includes text but also encompasses mediums such as speech and images, it is a natural idea to expand from uni-modal text-based large language models to multi-modal LLMs \citep{tang2023salmonn,huang2023language,zhang2023video,driess2023palm,moon2023anymal,chu2023qwen}. 
Most of this work focuses on enhancing the perceptual field of LLMs, enabling them to specifically understand auditory and visual capabilities. 
For example, LLaVA \citep{liu2023visual} combines a vision encoder and LLM into an end-to-end model for general-purpose visual-language understanding with impressive chat capabilities. Speech-LLaMA \citep{wu2023decoder} and SALMONN \citep{tang2023salmonn} try to perceive and understand all kinds of audio inputs with an additional audio encoder. 
Different from the above work, the goal of our work is to boost LLMs to generate speech instead of understanding speech.

Our work is also related to that large audio generative models \citep{borsos2022audiolm,wang2023neural_valle,zhang2023speak_vallex,rubenstein2023audiopalm,zhang2023speechgpt,chen2023lauragpt}. VALL-E \citep{chen2023lauragpt} is a novel and state-of-the-art zero-shot text-to-speech model, which contains an autogressive (AR) Transformer model and a non-autoregrressive (NAR) Transformer model to predict the first-layer quantized codes and rest-layer quantized codes separately. Our work follows the framework of VALL-E AR architecture to synthesize the speech with augmented LLMs. Besides, SpeechGPT \citep{zhang2023speechgpt} and AudioPaLM \citep{rubenstein2023audiopalm} convert speech into discrete hidden units and continually pre-train LLMs with hidden unit corpus. LauraGPT \citep{chen2023lauragpt} also fully fine-tunes LLMs with discrete codec codes of speech, to enable speech generation ability. However, no work has explored the use of existing speech synthesis models (e.g., VALL-E) to empower the speech generation capabilities of LLMs.
This paper focuses on empirically investigating and comparing different methods of endowing LLMs with speech synthesis capabilities.

\section{Methodology} \label{section_Method}

In this section, we will first introduce the core model components in the proposed framework in subsection \ref{model_components}, including large language model, speech compression model, and codec language model, then present the three integration strategies for LLMs and VALL-E in subsection \ref{integration_strategies}.

\subsection{Model Components}
\label{model_components}

There are three core components in our framework including a large language model (i.e., OPT \citep{zhang2022opt} or LLaMA \citep{touvron2023llama}), a speech compression model (i.e., Encodec \citep{defossez2022high}), and a codec language model (i.e., VALL-E \citep{wang2023neural_valle}). 
The large language model is employed to model textual tokens, with the option to include acoustic tokens as well. Meanwhile, the speech compression model is tasked with transforming continuous speech into discrete codec codes and subsequently reconstructing speech from these codes. Additionally, the codec language model is used to generate codec codes conditioning on the representation of textual tokens.


\paragraph{Large Language Model}
We conduct extensive experiments utilizing various pre-trained large language models including OPT \citep{zhang2022opt} models with different sizes including 125M, 350M, and 1.3B, and the LLaMA-7B \citep{touvron2023llama} model. These decoder-only models will be adapted using either full fine-tuning or parameter-efficient fine-tuning methods such as Low-rank Adaptation (LoRA) \citep{hu2021lora}. 
The OPT-125M/350M/1.3B model is a 12/24/24-layer Transformer decoder with an attention dimension of 768/1024/2048, respectively.
The LLaMA-7B model is a 32-layer Transformer decoder with an attention dimension of 4096.

\paragraph{Speech Compression Model}
To enable the LLM with speech generation ability, we utilize an external speech compression model EnCodec \citep{defossez2022high} to convert continuous speech into discrete codec codes.
EnCodec model is a convolution-based encoder-decoder network with residual vector quantization (RVQ) method.
It first tokenizes speech data into $L$-layer acoustic tokens using EnCodec encoder and RVQ module, and then recovers the speech waveform from all acoustic tokens using EnCodec decoder. In this paper, we adapt EnCodec with 6 kbps bandwidth and $L$=8 tokens for each frame.

\paragraph{Codec Language Model}

The neural codec language model VALL-E \citep{wang2023neural_valle} treats text-to-speech synthesis as a language model task, like GPT, and employs acoustic tokens (audio codec codes) as an intermediate representation of original speech. According to textual representations, VALL-E generates the codec code sequences (8 codes for each frame), from which final waveforms can be recovered by an audio compression decoder (e.g., Encodec). 
VALL-E contains two key modules, the auto-regressive (AR) codec language model and the non-autoregressive (NAR) codec language model. The former is responsible for predicting the acoustic tokens of the first codec code for each frame based on the semantic tokens in an auto-regressive manner, and the latter is used to generate the other 7-layer codes according to the sequence of the first-layer codes in parallel with the layer-level iterative generation method. 
In this work, we follow the VALL-E AR model, which is identical to the model architecture of LLMs, to augment LLMs with speech synthesis ability.

\subsection{Integration Strategies}
\label{integration_strategies}
We propose three methods to boost large language models with speech synthesis capability. Figure \ref{fig:methods} illustrates the different methods, including directly fine-tuned LLMs (Method A), superposed LLMs and VALL-E (Method B), and coupled LLMs and VALL-E (Method C).
Initially, we propose to directly fine-tune LLMs in Method A to determine if acoustic tokens can be integrated into LLMs by treating them as a novel language.
Furthermore, through Method B, we assess the capability of LLMs to encode both acoustic and textual tokens into a unified continuous embedding space, enhancing the performance of VALL-E in text-to-speech tasks. 
Finally, in Method C, we explore the potential of leveraging only the text encoding proficiency of LLMs to improve TTS outcomes without regarding acoustic tokens as a new language.

\paragraph{Method A: Directly Fine-tuned LLMs}
In order to verify whether acoustic tokens can be incorporated into LLMs by simply regarding it as a new language, enabling the joint training of both acoustic and textual tokens, the most straightforward approach involves fine-tuning language models directly with TTS training data by either full fine-tuning or parameter-efficient fine-tuning, as shown in Figure \ref{fig:methods}(\subref{fig:methodA}). 
Through training on TTS data, we also augment large language models with speech synthesis ability at the same time.
In practice, we found that using parameter-efficient fine-tuning methods such as LoRA in this way is less effective and results in relatively poor performance. We speculate that this is because large language models do not have the ability to generate codec codes inherently and it is more difficult for LLMs to generate speech than understand speech signals. 
Therefore, we directly fully fine-tune LLMs as one kind of approach that endows LLMs with speech synthesis ability.

\paragraph{Method B: Superposed LLMs and VALL-E} 
Inspired by the observation of Method A introduced above, we aim to further explore the suitability of LLMs for encoding both acoustic tokens and textual tokens into continuous embedding space so that this representation can be used by VALL-E to perform TTS tasks better.
As shown in Figure \ref{fig:methods}(\subref{fig:methodB}), in this approach, we superpose the pre-trained LLMs and VALL-E models to promote the speech generation ability of LLMs. Both textual tokens and acoustic tokens are encoded by LLM, and are sent to the codec language model to predict the first-layer codec code. Besides, a linear projection layer is added between LLM and codec language model to bridge the dimension gap between them.

\paragraph{Method C: Coupled LLMs and VALL-E}
Given the distinct roles and strengths of LLMs and VALL-E, it would be interesting to investigate the effect of only utilizing the text encoding ability of LLMs, instead of treating acoustic tokens as a new language in previous methods, to promote TTS performance of VALL-E.
Therefore, another natural idea is to take full use of the advantages of LLMs and VALL-E, and cascade the pre-trained LLMs and VALL-E into an end-to-end model.
LLMs excel at encoding and generating text, while VALL-E specializes in producing speech tokens based on textual tokens.
Hence, in this text-to-speech framework, we first use LLMs to encode text and get better text representation, then feed it to VALL-E as text input, as shown in Figure \ref{fig:methods}(\subref{fig:methodC}). In this method, we also incorporate a linear projection layer between the LLM and the codec language model to reconcile the disparity in dimensions.

\section{Experiments} \label{section:experiment}

\subsection{Experiment Setup}

\paragraph{Dataset:}
Pre-trained models are fine-tuned on two ASR datasets, which can also be used to train TTS tasks as VALL-E (X) \citep{wang2023neural_valle, zhang2023speak_vallex}. Specifically, we use LibriSpeech (LS, 960 hours) \citep{7178964_librispeech960dataset} and the English part of Multilingual LibriSpeech (MLS) \citep{pratap20_interspeech_mlsdataset}\footnote{We do not use Librilight \citep{Kahn2020LibriLight} data like VALL-E, due to its lack of ground-truth transcriptions required for tokenization using large language model's tokenizer.}. The Multilingual LibriSpeech is a 50K-hour ASR corpus including 8 languages derived from read audiobooks of LibriVox, where English accounts for about 44.5K hours predominately.
We evaluate our proposed methods on the LibriSpeech dev-clean, dev-other, test-clean, and test-other datasets. 
We use the samples that range in duration from 4 to 20 seconds from these datasets\footnote{Note that VALL-E (X)'s evaluation set contains audio samples ranging from 4 to 10 seconds in length. Given that the audio durations within the MLS dataset span 10 to 20 seconds, our model demonstrates the capability to perform speech synthesis tasks over extended periods.}. 
Following \cite{wang2023neural_valle}, we use the first 3 seconds of the ground-truth speech as prompts for each sample synthesis. Each experiment is conducted thrice, with the average score being reported.

\paragraph{Data Preprocessing:}
To unify the training of speech and text modalities, we transform both into discrete tokens.
In our approach, ASR data transcriptions are tokenized into subwords (semantic tokens) with the tokenizer from large language models. Meanwhile, speech data are quantized into acoustic tokens using the EnCodec, which operates at a 6 kbps bandwidth and a downsampling ratio of 320, producing 8 acoustic tokens per frame and 75 frames per second of audio. We concatenate the semantic tokens and corresponding acoustic tokens to form a cohesive training sample.


\subsection{Training Details}
For Method A, we employ both LoRA and full fine-tuning techniques to train OPT models. However, due to computational resource limitations, we exclusively utilize LoRA for training the LLaMA-7B model. Additionally, we augment the LLMs' vocabulary with acoustic tokens, specifically incorporating 1024 Encodec tokens in our configuration.
In Method B, we introduce LoRA parameters to LLM and codec language model respectively. The LLM is initialized with either a pre-trained OPT-350M or LLaMA-7B, while the codec language model is initialized with a pre-trained VALL-E. We also expand the vocabulary of LLM with acoustic tokens like Method A. Besides, the input acoustic and textual embeddings from VALL-E are omitted, as the LLM now provides the representations for both acoustic and textual tokens.
Similarly, in Method C we also add LoRA parameters to pre-trained LLM and pre-trained VALL-E respectively, and discard the textual token embedding of VALL-E.
We fix the LoRA parameter to $R = 64$ for adjusting self-attention parameters. 
Consequently, using Method A for LoRA training yields approximately 14M trainable parameters for OPT-350M and 71M for LLaMA-7B. In contrast, Method B incorporates codec code embedding, LoRA, and linear projection, resulting in around 21M trainable parameters for OPT-350M and 82M for LLaMA-7B. Meanwhile, Method C reduces the count of trainable parameters to 20M for OPT-350M and 78M for LLaMA-7B, as it does not utilize codec code embedding for the LLMs.
Our models are trained using the Adam optimizer with $\beta_1 = 0.9$ and $\beta_2 = 0.98$ \citep{DBLP:KingmaB14_iclr_adam}. All models are trained on TTS tasks for 400K steps on 32 V100 GPUs with a batch size of 100 seconds per GPU. The maximum learning rate is $5 \times 10^{-4}$ with a warm-up step of 40K. We follow the configuration of VALL-E to train our non-autoregressive language model as introduced in Section \ref{model_components}.

\subsection{Evaluation Metrics}
\label{sec:evaluation_metrics}
We use the automatic evaluation metrics, including the word error rate (WER), speaker similarity (SS), and speech naturalness (SN) to evaluate the generated speech for simplicity and convenience. The WER score is obtained by an open-source Conformer Transducer model\footnote{\url{https://github.com/NVIDIA/NeMo/}}, ranging from 0 to 100. The lower the WER, the more accurate the generated speech is. Given generated and prompt speech utterances, the SS is measured by an automatic speaker verification (ASV) WavLM \citep{chen2022wavlm} model\footnote{\url{https://github.com/microsoft/UniSpeech/tree/main/downstreams/speaker_verification}}, ranging from -1 to 1. The larger the SS, the more similar the speakers of the two utterances are. SN score of generated speech is measured by the open-source NISQA\footnote{\url{https://github.com/gabrielmittag/NISQA}} \citep{mittag20_interspeech_nisqa}. Since we mainly use LoRA to fine-tune LLMs, the original textual processing ability of LLMs will not be affected when performing NLP tasks without LoRA parameters, therefore NLP tasks are not evaluated in this paper.

\subsection{Inference Strategies}
\label{sec:inference_strategies}
After training, we use sampling methods for our models to generate the acoustic tokens of the first layer codec codes. Specifically, we use top-$p$ \citep{holtzman2019curious} sampling with $p=1.0$ and temperature is 1.0. We adopt three different strategies to choose sampled sequences following previous work \citep{wang2023viola}.
\begin{itemize}
    \item Strategy \Rmnum{1} performs only one synthesis inference for one text, and then the sampled acoustic sequence is chosen as the final result.
    \item Strategy \Rmnum{2} conducts five synthesis inferences for a single text, selecting the utterance that yields the highest speaker similarity score.
    \item Strategy \Rmnum{3} also performs five synthesis inferences for a given text and selects the utterance that exhibits the lowest word error rate.
\end{itemize}

\subsection{Main Results}
\label{subsec:main_result}
We synthesize the English speech of corresponding text prompted by a 3s English speech utterance on selected samples of dev-clean, dev-other, test-clean, and test-other datasets, where Table \ref{tab:main_result} shows the results of dev-clean and others are shown in Appendix \ref{sec:appendix_main_result}. 
As summarized in Table \ref{tab:main_result}, we replicate the VALL-E baseline using parameters identical to those of \cite{wang2023neural_valle}, while the proposed three methods are validated using both LLaMA-7B and OPT-350M models. 
We apply the three inference strategies outlined in Section \ref{sec:inference_strategies}, evaluating their performance using the metrics of word error rate (WER), sentence similarity (SS), and speaker naturalness (SN), as introduced in Section \ref{sec:evaluation_metrics}.

According to the experimental results, we can draw three conclusions:
(1) Directly fine-tuning LLMs by LoRA performs worse than the VALL-E baseline model. Although full fine-tuning can mitigate the problem and achieve comparable performance with VALL-E, it needs massive computational resources for large models.
(2) Method B, when employed with both the OPT-350M or LLaMA-7B models, surpasses the VALL-E baseline in terms of WER, SS, and SN, which demonstrates that augmenting LLM with VALL-E can address the above challenge with LoRA methods, given that LLMs are capable of encoding both acoustic and textual tokens and VALL-E shares a portion of the burden for speech synthesis in LLMs.
(3) By fully leveraging the respective strengths of both components, Method C achieves the best performance among the proposed methods, which significantly outperforms VALL-E on word error rate, speaker similarity, and speech naturalness. 
Compared to the VALL-E, the word error rate of Method C with LLaMA-7B is relatively decreased by 10.9\%, 14.3\%, and 6.9\% under inference Strategy \Rmnum{1}, \Rmnum{2}, and \Rmnum{3} respectively, the speaker similarity is relatively improved by 0.02, 0.03, and 0.03, and the speech naturalness is improved by 0.03, 0.02, and 0.02 respectively.

\begin{table*}[!htp]
\centering
\footnotesize
\renewcommand\arraystretch{1.2}
\begin{adjustbox}{width=0.99\textwidth,center}
\begin{tabular}{ccccccccccc} \toprule
 \multirow{2}*{\textbf{Methods}} & \multirow{2}*{\textbf{LLMs}}  & \multicolumn{3}{c}{\textbf{Strategy \Rmnum{1}}} & \multicolumn{3}{c}{\textbf{Strategy \Rmnum{2}}} & \multicolumn{3}{c}{\textbf{Strategy \Rmnum{3}}}   \\ 
~ & ~ & \textbf{WER}$\downarrow$ & \textbf{SS}$\uparrow$ & \textbf{SN}$\uparrow$ & \textbf{WER}$\downarrow$ & \textbf{SS}$\uparrow$ & \textbf{SN}$\uparrow$ & \textbf{WER}$\downarrow$ & \textbf{SS}$\uparrow$ & \textbf{SN}$\uparrow$  \\
\midrule
VALL-E & - & 4.39 & 0.52 & 3.26 & 4.27 & 0.58 & 3.28 & 1.31 &  0.56  & 3.27 \\
\hdashline
\multirow{3}*{A} & OPT-350M & 10.28 & 0.49 & 3.20 & 9.74 & 0.53 & 3.21 & 3.97 & 0.51 & 3.20  \\
 & OPT-350M FT$^*$  & 4.21 & 0.53 & 3.28 & 4.08 & 0.60 & 3.29 & 1.28 &  0.58 & 3.28 \\
& LLaMA-7B  & 9.61 & 0.49 & 3.20 & 9.19 & 0.54 & 3.21 & 3.63 & 0.51 & 3.21 \\
\hdashline
\multirow{2}*{B}  & OPT-350M  & 4.12 & 0.53 & 3.28 & 3.94 & \textbf{0.61} & 3.29 & 1.25 & 0.57 & 3.29 \\
& LLaMA-7B  & 4.05 & 0.53 & 3.29 & 3.82 & \textbf{0.61} & \textbf{3.30} & 1.23 & 0.58 & 3.29 \\
\hdashline
\multirow{2}*{C} & OPT-350M  & 3.99 & \textbf{0.54} & \textbf{3.30} & 3.72 & \textbf{0.61} & 3.29 & 1.26 & \textbf{0.59}  & \textbf{3.30}    \\
 &  LLaMA-7B  & \textbf{3.91} & \textbf{0.54} & 3.29 & \textbf{3.66} & \textbf{0.61} & \textbf{3.30} & \textbf{1.22} &  \textbf{0.59} & 3.29    \\
\bottomrule
\end{tabular}
\end{adjustbox}
\caption{Main evaluation results on LibriSpeech dev-clean dataset. FT$^*$ means full fine-tuning, and other models adopt LoRA techniques. VALL-E is the text-to-speech baseline, Method A/B/C are introduced in Section \ref{integration_strategies}, and inference strategies \Rmnum{1}/\Rmnum{2}/\Rmnum{3} are listed in Section \ref{sec:inference_strategies}.
}
\label{tab:main_result}
\end{table*}

\subsection{Analysis}
To facilitate a clearer comprehension of our method, we conduct detailed analyses and ablation studies in this section.

\begin{figure*}[!htp]
  \centering
  \includegraphics[width=1.0\textwidth]{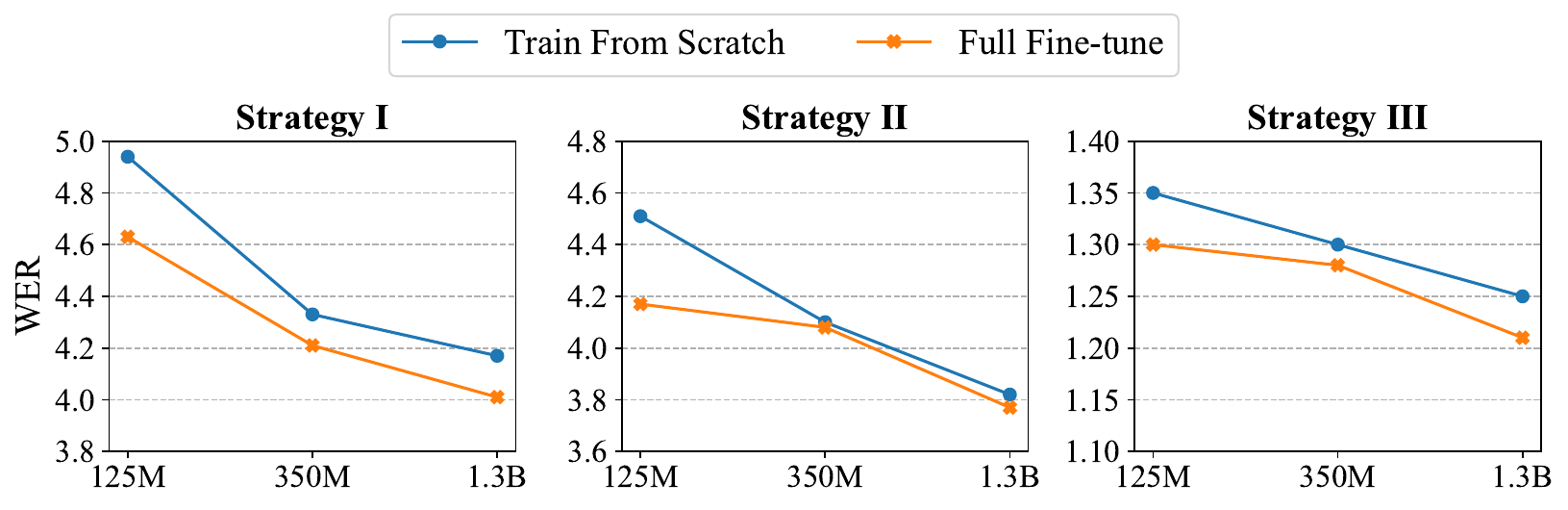}
  \caption{WER results of using different model sizes in Method A under three inference strategies introduced in Section \ref{sec:inference_strategies}. The overall results including speaker similarity and speech naturalness are summarized in Appendix \ref{sec:appendix_effect_model_size}.
  }
  \label{fig:analysis_model_size}
\end{figure*}

\paragraph{Effect of Model Size}
The capacity of a large language model is significantly influenced by its parameter number. Consequently, we explore the impact of varying model sizes within the OPT framework through direct full fine-tuning (referred to as Method A in Table \ref{tab:main_result}), examining models with 125M, 350M, and 1.3B parameters.
Additionally, we establish baselines by training these models from scratch. We conduct this experiment on the dev-clean dataset, the results of which are depicted in Figure \ref{fig:analysis_model_size}.
The comparison between the two curves illustrates the effectiveness of using pre-trained LLMs. The largest OPT model with 1.3B parameters achieves the best performance overall compared to 125M and 350M. 
This finding suggests that increasing the model size could be a viable strategy for enhancing speech synthesis capabilities.

\paragraph{Effect of Continual Pre-training}
Since unlabeled speech data is more common than paired speech-text data, we also investigate the way of taking advantage of massive unlabeled speech data to promote the speech synthesis performance of LLMs. 
Specifically, inspired by the next token prediction pre-training objective of decoder-only language models like GPT, we use EnCodec codes of the LibriLight \citep{Kahn2020LibriLight} dataset to continually pre-train large language models, so that they can adapt to speech modality better. 
Then we use paired speech-text data to fine-tune continually pre-trained models and compare them with those that have not been continually pre-trained.
Table \ref{tab:analysis_continue_pretrain_mls} shows the comparison results of (1) training from scratch, (2) directly full fine-tuning, and (3) continually pre-training and then full fine-tuning, on large (MLS+LS) and small (LS) datasets. 
The experimental results on Method A with OPT-350M as LLM show that the continual pre-training method achieves significant WER reduction than methods of full fine-tuning and training from scratch on the small fine-tuning dataset, and obtains comparable performance on the large fine-tuning dataset.

\begin{table*}[!htp]
\centering
\footnotesize
\renewcommand\arraystretch{1.2}
\begin{adjustbox}{width=0.99\textwidth,center}
\begin{tabular}{ccccccccccc} \toprule
\multirow{2}*{\textbf{Data}} & \multirow{2}*{\textbf{Method}}  & \multicolumn{3}{c}{\textbf{Strategy \Rmnum{1}}} & \multicolumn{3}{c}{\textbf{Strategy \Rmnum{2}}} & \multicolumn{3}{c}{\textbf{Strategy \Rmnum{3}}}   \\ 
~ & ~ & \textbf{WER}$\downarrow$ & \textbf{SS}$\uparrow$ & \textbf{SN}$\uparrow$ & \textbf{WER}$\downarrow$ & \textbf{SS}$\uparrow$ & \textbf{SN}$\uparrow$ & \textbf{WER}$\downarrow$ & \textbf{SS}$\uparrow$ & \textbf{SN}$\uparrow$  \\
\midrule
\multirow{3}*{MLS+LS} & Train From Scratch & 4.33 & 0.52 & 3.26 & 4.10 & 0.59 & 3.28 & 1.30 & 0.56 & 3.27 \\
& Full Fine-tune  & 4.21 & 0.53 & 3.28 & 4.08 & 0.60 & 3.29 & 1.28 &  0.58 & 3.28 \\
& Pre-train+Fine-tune & 4.19 & 0.53 & 3.28 & 4.03 & 0.60 & 3.29 & 1.26 & 0.58 & 3.28 \\
\hdashline
\multirow{3}*{LS} & Train From Scratch & 5.71 & 0.51 & 3.26 & 5.11 & 0.58 & 3.28 & 1.97 & 0.55 & 3.28 \\
 & Full Fine-tune  & 5.65 & 0.50 & 3.26 & 5.10 & 0.57 & 3.27 & 1.99 & 0.53 & 3.28 \\
 & Pre-train+Fine-tune & 5.47 & 0.51 & 3.26 & 4.99 & 0.58 & 3.29 & 1.91 & 0.55 & 3.30 \\
\bottomrule
\end{tabular}
\end{adjustbox}
\caption{Effect of continual pre-training on dev-clean set with Method A and OPT-350M. MLS+LS means that the fine-tuning data are Multilingual LibriSpeech and LibriSpeech, and LS means Librispeech only.}
\label{tab:analysis_continue_pretrain_mls}
\end{table*}


\paragraph{Effect of Pre-trained VALL-E}
To validate the benefits of employing the pre-trained codec language model VALL-E, we undertake an ablation study focusing on the impact of random initialization versus pre-trained initialization. Specifically, we fully fine-tune the randomly initialized VALL-E but use LoRA to fine-tune the VALL-E initialized with pre-trained weights. Table \ref{tab:analysis_pretrained_valle} delineates the performance disparity between models with Method B that begin with random weights and those initialized with pre-trained VALL-E. The results clearly indicate that initializing with pre-trained VALL-E results in fewer trainable parameters and significantly surpasses random initialization across various inference strategies and evaluation criteria.

\begin{table*}[!htp]
\centering
\footnotesize
\renewcommand\arraystretch{1.2}
\begin{adjustbox}{width=0.99\textwidth,center}
\begin{tabular}{ccccccccccc} \toprule 
\multirow{2}*{\textbf{LLMs}} & \multirow{2}*{\textbf{VALL-E}}  & \multicolumn{3}{c}{\textbf{Strategy \Rmnum{1}}} & \multicolumn{3}{c}{\textbf{Strategy \Rmnum{2}}} & \multicolumn{3}{c}{\textbf{Strategy \Rmnum{3}}}   \\ 
~ & ~ & \textbf{WER}$\downarrow$ & \textbf{SS}$\uparrow$ & \textbf{SN}$\uparrow$ & \textbf{WER}$\downarrow$ & \textbf{SS}$\uparrow$ & \textbf{SN}$\uparrow$ & \textbf{WER}$\downarrow$ & \textbf{SS}$\uparrow$ & \textbf{SN}$\uparrow$  \\
\midrule
\multirow{2}*{OPT-350M} & Randomly (FT$^*$) & 4.31 & 0.52 & 3.27 & 4.09 & 0.59 & 3.28 & 1.36 & 0.56 & 3.27    \\
 & Pre-trained & 4.12 & 0.53 & 3.28 & 3.94 & 0.61 & 3.29 & 1.25 & 0.57 & 3.29 \\
\multirow{2}*{LLaMA-7B} & Randomly (FT$^*$) & 4.27 & 0.52 & 3.27 & 4.11  & 0.59 & 3.28 & 1.32 & 0.56 & 3.28 \\
 & Pre-trained & 4.05 & 0.53 & 3.29 & 3.82 & 0.61 & 3.30 & 1.23 & 0.58 & 3.29 \\
\bottomrule
\end{tabular}
\end{adjustbox}
\caption{Effect of pre-trained VALL-E on dev-clean set with method B, where VALL-E is either randomly initialized or is leveraged as a pre-trained model. FT$^*$ means full fine-tuning, and models with pre-trained VALL-E adopt LoRA techniques.}
\label{tab:analysis_pretrained_valle}
\end{table*}

\paragraph{LoRA vs. Full Fine-tuning in VALL-E}
The previous section has demonstrated that pre-trained VALL-E enhanced with LoRA outperforms a randomly initialized version of VALL-E.
Besides, the main results also indicate that fully fine-tuning OPT-350M yields better results than applying LoRA techniques.
Since the model size of VALL-E is relatively small compared to that of LLMs, we are now keen to investigate the peak performance achievable by substituting LoRA with full fine-tuning in VALL-E.
Table \ref{tab:analysis_lora_and_full_finetune} presents a comparison of performance between LoRA fine-tuning and full fine-tuning approaches for VALL-E, revealing that full fine-tuning can indeed lead to further enhancements in performance.

\begin{table*}[!htp]
\centering
\footnotesize
\renewcommand\arraystretch{1.2}
\begin{adjustbox}{width=0.99\textwidth,center}
\begin{tabular}{ccccccccccc} \toprule 
\multirow{2}*{\textbf{LLMs}} & \multirow{2}*{\textbf{VALL-E}}  & \multicolumn{3}{c}{\textbf{Strategy \Rmnum{1}}} & \multicolumn{3}{c}{\textbf{Strategy \Rmnum{2}}} & \multicolumn{3}{c}{\textbf{Strategy \Rmnum{3}}}   \\ 
~ & ~ & \textbf{WER}$\downarrow$ & \textbf{SS}$\uparrow$ & \textbf{SN}$\uparrow$ & \textbf{WER}$\downarrow$ & \textbf{SS}$\uparrow$ & \textbf{SN}$\uparrow$ & \textbf{WER}$\downarrow$ & \textbf{SS}$\uparrow$ & \textbf{SN}$\uparrow$  \\
\midrule
\multirow{2}*{OPT-350M} & LoRA & 3.99 & 0.54 & 3.30 & 3.72 & 0.61 & 3.29 & 1.26 & 0.59 & 3.30  \\
 & Full Fine-tune  & 3.97 & 0.54 & 3.31 & 3.64 & 0.61 & 3.30 & 1.25 & 0.59 & 3.31 \\
\multirow{2}*{LLaMA-7B} & LoRA & 3.91 & 0.54 & 3.29 & 3.66 & 0.61 & 3.30 & 1.22 & 0.59 & 3.29    \\
 & Full Fine-tune  & 3.90 & 0.54 & 3.31 & 3.46 & 0.61 & 3.31 & 1.20 & 0.59 & 3.31    \\
\bottomrule
\end{tabular}
\end{adjustbox}
\caption{Comparison of LoRA and full fine-tuning of VALL-E on dev-clean set with Method C.}
\label{tab:analysis_lora_and_full_finetune}
\end{table*}

\section{Conclusion}
\label{conclusion}
In this study, we explore various strategies for incorporating speech synthesis capabilities into large language models (LLMs).
Our findings show that simply fine-tuning LLMs with LoRA fails to match the performance of the baseline, indicating the challenge of enhancing LLMs with speech synthesis capabilities.
Further investigation demonstrates that LLMs augmented with a pre-trained text-to-speech synthesis model can surpass the performance of the baseline VALL-E model.
In particular, by leveraging the respective strengths of LLMs and VALL-E, the coupled LLM and VALL-E method achieves the highest performance among the methods evaluated.
Moreover, we conduct comprehensive analyses to better understand the proposed LLMs augmented with speech synthesis ability.



\bibliography{neurips_2023}
\bibliographystyle{plainnat}

\input{appendix.tex}

\end{document}

%% file: appendix.tex
\appendix

\section{Main Results of dev-other, test-clean, test-other}
\label{sec:appendix_main_result}
Throughout this section, we list the main results of dev-other, test-clean, and test-other in Table \ref{tab:appendix_main_result_dev_other}, \ref{tab:appendix_main_result_test_clean}, and \ref{tab:appendix_main_result_test_other} respectively.
All these three tables show the same trend as in Table \ref{tab:main_result}, which can further consolidate the conclusions summarized in Section \ref{subsec:main_result}.

\begin{table*}[!htp]
\centering
\footnotesize
\renewcommand\arraystretch{1.2}
\begin{adjustbox}{width=0.99\textwidth,center}
\begin{tabular}{ccccccccccc} \toprule 
 \multirow{2}*{\textbf{Methods}} & \multirow{2}*{\textbf{LLMs}}  & \multicolumn{3}{c}{\textbf{Strategy \Rmnum{1}}} & \multicolumn{3}{c}{\textbf{Strategy \Rmnum{2}}} & \multicolumn{3}{c}{\textbf{Strategy \Rmnum{3}}}   \\ 
~ & ~ & \textbf{WER}$\downarrow$ & \textbf{SS}$\uparrow$ & \textbf{SN}$\uparrow$ & \textbf{WER}$\downarrow$ & \textbf{SS}$\uparrow$ & \textbf{SN}$\uparrow$ & \textbf{WER}$\downarrow$ & \textbf{SS}$\uparrow$ & \textbf{SN}$\uparrow$  \\
\midrule
VALL-E & - & 7.65 & 0.47 & 3.07 & 7.44 & 0.55 & 3.08 & 2.53 &  0.53  & 3.08 \\
\hdashline
\multirow{3}*{A} & OPT-350M & 14.76 & 0.41 & 2.97 & 13.93 & 0.50 & 3.00 & 4.48 & 0.47 & 3.03 \\
 & OPT-350M FT$^*$  & 7.39 & 0.47 & 3.08 & 7.00 & 0.56 & 3.09 & 2.45 & 0.54 & 3.08 \\
& LLaMA-7B & 14.43 & 0.41 & 2.98 & 13.21 & 0.51 & 3.00 & 4.41 & 0.46 & 3.02 \\
\hdashline
\multirow{2}*{B}  & OPT-350M  & 6.90 & 0.49 & 3.08 & 6.76 & 0.57 & 3.09 & 2.35 & 0.54 & 3.08 \\
& LLaMA-7B  & 6.99 & 0.49 & 3.08 & 6.81 & \textbf{0.58} & 3.08 & \textbf{2.31} & \textbf{0.55} & 3.08 \\
\hdashline
\multirow{2}*{C} & OPT-350M  & 6.94 & \textbf{0.50} & 3.08 & 6.75 & 0.57 & 3.09 & 2.33 & 0.54  & \textbf{3.10}    \\
 &  LLaMA-7B  & \textbf{6.85} & \textbf{0.50} & \textbf{3.09} & \textbf{6.60} & \textbf{0.58} & \textbf{3.10} & \textbf{2.31} &  0.54 & \textbf{3.10}    \\
\bottomrule
\end{tabular}
\end{adjustbox}
\caption{Main evaluation results on LibriSpeech dev-other dataset. FT$^*$ means full fine-tuning, and the other models adapt LoRA techniques. VALL-E is the text-to-speech baseline, Method A/B/C are introduced in Section \ref{integration_strategies}, and inference strategies \Rmnum{1}/\Rmnum{2}/\Rmnum{3} are listed in Section \ref{sec:inference_strategies}.
}
\label{tab:appendix_main_result_dev_other}
\end{table*}

\begin{table*}[!htp]
\centering
\footnotesize
\renewcommand\arraystretch{1.2}
\begin{adjustbox}{width=0.99\textwidth,center}
\begin{tabular}{ccccccccccc} \toprule 
 \multirow{2}*{\textbf{Methods}} & \multirow{2}*{\textbf{LLMs}}  & \multicolumn{3}{c}{\textbf{Strategy \Rmnum{1}}} & \multicolumn{3}{c}{\textbf{Strategy \Rmnum{2}}} & \multicolumn{3}{c}{\textbf{Strategy \Rmnum{3}}}   \\ 
~ & ~ & \textbf{WER}$\downarrow$ & \textbf{SS}$\uparrow$ & \textbf{SN}$\uparrow$ & \textbf{WER}$\downarrow$ & \textbf{SS}$\uparrow$ & \textbf{SN}$\uparrow$ & \textbf{WER}$\downarrow$ & \textbf{SS}$\uparrow$ & \textbf{SN}$\uparrow$  \\
\midrule
VALL-E & - & 4.52 & 0.51 & 3.31 & 4.33 & 0.58 & 3.31 & 1.31 &  0.56  & 3.31 \\
\hdashline
\multirow{3}*{A} & OPT-350M & 10.66 & 0.47 & 3.21 & 10.01 & 0.54 & 3.21 & 4.01 & 0.51 & 3.21 \\
 & OPT-350M FT$^*$  & 4.26 & 0.51 & 3.31 & 3.98 & 0.59 & 3.31 & 1.23 & 0.57 & 3.31 \\
& LLaMA-7B & 10.49 & 0.47 & 3.20 & 9.72 & 0.55 & 3.21 & 3.95 & 0.52 & 3.22 \\
\hdashline
\multirow{2}*{B}  & OPT-350M  & 3.91 & \textbf{0.53} & 3.30 & 3.54 & \textbf{0.60} & 3.31 & 1.17 & 0.57 & 3.31 \\
& LLaMA-7B  & 3.86 & \textbf{0.53} & 3.30 & 3.44 & \textbf{0.60} & 3.32 & \textbf{1.16} & 0.57 & 3.31 \\
\hdashline
\multirow{2}*{C} & OPT-350M  & 3.89 & 0.52 & \textbf{3.32} & 3.54 & \textbf{0.60} & \textbf{3.33} & 1.25 & \textbf{0.58}  & \textbf{3.34}    \\
 &  LLaMA-7B  & \textbf{3.71} & \textbf{0.53} & 3.31 & \textbf{3.43} & \textbf{0.60} & 3.32 & \textbf{1.16} &  \textbf{0.58} & 3.32    \\
\bottomrule
\end{tabular}
\end{adjustbox}
\caption{Main evaluation results on LibriSpeech test-clean dataset. FT$^*$ means full fine-tuning, and the other models adapt LoRA techniques. VALL-E is the text-to-speech baseline, Method A/B/C are introduced in Section \ref{integration_strategies}, and inference strategies \Rmnum{1}/\Rmnum{2}/\Rmnum{3} are listed in Section \ref{sec:inference_strategies}.
}
\label{tab:appendix_main_result_test_clean}
\end{table*}

\begin{table*}[!htp]
\centering
\footnotesize
\renewcommand\arraystretch{1.2}
\begin{adjustbox}{width=0.99\textwidth,center}
\begin{tabular}{ccccccccccc} \toprule 
 \multirow{2}*{\textbf{Methods}} & \multirow{2}*{\textbf{LLMs}}  & \multicolumn{3}{c}{\textbf{Strategy \Rmnum{1}}} & \multicolumn{3}{c}{\textbf{Strategy \Rmnum{2}}} & \multicolumn{3}{c}{\textbf{Strategy \Rmnum{3}}}   \\ 
~ & ~ & \textbf{WER}$\downarrow$ & \textbf{SS}$\uparrow$ & \textbf{SN}$\uparrow$ & \textbf{WER}$\downarrow$ & \textbf{SS}$\uparrow$ & \textbf{SN}$\uparrow$ & \textbf{WER}$\downarrow$ & \textbf{SS}$\uparrow$ & \textbf{SN}$\uparrow$  \\
\midrule
VALL-E & - & 8.45 & 0.46 & 3.08 & 8.09 & 0.54 & 3.09 & 2.98 & 0.50  & 3.09 \\
\hdashline
\multirow{3}*{A} & OPT-350M & 14.42 & 0.41 & 2.99 & 13.79 & 0.51 & 3.01 & 4.51 & 0.47 & 3.03 \\
 & OPT-350M FT$^*$  & 8.23 & 0.46 & 3.09 & 7.86 & 0.54 & 3.10 & 2.86 & 0.50 & 3.10 \\
& LLaMA-7B & 14.17 & 0.42 & 3.00 & 13.59 & 0.51 & 3.02 & 4.36 & 0.47 & 3.03 \\
\hdashline
\multirow{2}*{B}  & OPT-350M  & 7.68 & 0.47 & \textbf{3.10} & 7.53 & 0.55 & 3.10 & 2.79 & 0.50 & 3.10 \\
& LLaMA-7B  & 7.67 & 0.47 & \textbf{3.10} & 7.40 & 0.55 & 3.10 & 2.76 & 0.51 & \textbf{3.11} \\
\hdashline
\multirow{2}*{C} & OPT-350M  & 7.78 & 0.47 & \textbf{3.10} & 7.42 & \textbf{0.56} & 3.10 & 2.76 & 0.51  & \textbf{3.11}    \\
 &  LLaMA-7B  & \textbf{7.62} & \textbf{0.48} & \textbf{3.10} & \textbf{7.14} & \textbf{0.56} & \textbf{3.11} & \textbf{2.71} &  \textbf{0.52} & \textbf{3.11}    \\
\bottomrule
\end{tabular}
\end{adjustbox}
\caption{Main evaluation results on LibriSpeech test-other dataset. FT$^*$ means full fine-tuning, and the other models adapt LoRA techniques. VALL-E is the text-to-speech baseline, Method A/B/C are introduced in Section \ref{integration_strategies}, and inference strategies \Rmnum{1}/\Rmnum{2}/\Rmnum{3} are listed in Section \ref{sec:inference_strategies}.
}
\label{tab:appendix_main_result_test_other}
\end{table*}

\section{Effect of Model Size: Detailed Results}
\label{sec:appendix_effect_model_size}
Table \ref{tab:appendix_analysis_model_size} shows the detailed word error rate, speaker similarity, and speech naturalness results of using different model sizes in Method A under three inference strategies introduced in Section \ref{sec:inference_strategies}. 


\begin{table*}[!htp]
\centering
\footnotesize
\renewcommand\arraystretch{1.2}
\begin{adjustbox}{width=0.99\textwidth,center}
\begin{tabular}{ccccccccccc} \toprule 
\multirow{2}*{\textbf{Methods}} & \multirow{2}*{\textbf{Model}}  & \multicolumn{3}{c}{\textbf{Strategy \Rmnum{1}}} & \multicolumn{3}{c}{\textbf{Strategy \Rmnum{2}}} & \multicolumn{3}{c}{\textbf{Strategy \Rmnum{3}}}   \\ 
~ & ~ & \textbf{WER}$\downarrow$ & \textbf{SS}$\uparrow$ & \textbf{SN}$\uparrow$ & \textbf{WER}$\downarrow$ & \textbf{SS}$\uparrow$ & \textbf{SN}$\uparrow$ & \textbf{WER}$\downarrow$ & \textbf{SS}$\uparrow$ & \textbf{SN}$\uparrow$  \\
\midrule
\multirow{3}*{Train From Scratch} & OPT-125M & 4.94 & 0.51 & 3.26 & 4.51 & 0.58 & 3.26 & 1.35 & 0.56 & 3.27 \\
~ & OPT-350M & 4.33 & 0.52 & 3.26 & 4.10 & 0.59 & 3.28 & 1.30 & 0.56 & 3.27 \\
~ & OPT-1.3B& 4.17 & 0.52 & 3.27 & 3.82 & 0.59 & 3.27 & 1.25 & 0.58 & 3.28 \\
\hdashline
\multirow{3}*{Full Fine-tune} & OPT-125M  & 4.63 & 0.53 & 3.26 & 4.17 & 0.60 & 3.28 & 1.30 & 0.58 & 3.28 \\
~ & OPT-350M & 4.21 & 0.53 & 3.28 & 4.08 & 0.60 & 3.29 & 1.28 &  0.58 & 3.28 \\
~ & OPT-1.3B & 4.01 & 0.53 & 3.28 & 3.77 & 0.60 & 3.29 & 1.21 & 0.59 & 3.30 \\
\bottomrule
\end{tabular}
\end{adjustbox}
\caption{WER, SS, and SN results of using different model sizes in Method A under three inference strategies.}
\label{tab:appendix_analysis_model_size}
\end{table*}